# Logic Contrastive Reasoning with Lightweight Large Language Model for Math Word Problems

Ding Kai[1], Ma Zhenguo[1], Yan Xiaoran[1*]
[1]Research Center for Scientific data open and sharing hub, Zhejiang Lab, Zhejiang Province, 311121 P.R. China

**Abstract：**

This study focuses on improving the performance of lightweight Large Language Models (LLMs) in mathematical reasoning tasks. We introduce a novel method for measuring mathematical logic similarity and design an automatic screening mechanism to construct a set of reference problems that integrate both semantic and logical similarity. By employing carefully crafted positive and negative example prompts, we guide the model towards adopting sound reasoning logic. To the best of our knowledge, this is the first attempt to utilize retrieval-enhanced generation for mathematical problem-solving. Experimental results demonstrate that our method achieves a 15.8% improvement over the Chain of Thought approach on the SVAMP dataset and a 21.5 % improvement on the GSM8K dataset. Further application of this method to a large-scale model with 175 billion parameters yields performance comparable to the best results on both aforementioned datasets. Finally, we conduct an analysis of errors during the reasoning process, providing valuable insights and directions for future research on reasoning tasks using large language models. Code and dataset are available at: https://github.com/derby-ding/llm-math-reasoning/tree/main

## 1. Introduction

Recent years have witnessed remarkable advancements in Large Language Models (LLMs) [1], achieving state-of-the-art (SOTA) performance across diverse tasks and datasets. Despite their widespread deployment in real-world applications, LLMs continue to grapple with critical challenges, most notably the "hallucination" problem. This issue is particularly pronounced in tasks demanding precise logical reasoning, such as mathematical problem-solving and commonsense inference, significantly impacting the reliability and practical utility of these models.

The prevalence of hallucinations has ignited a vigorous debate within the research community regarding the true extent of LLMs' logical reasoning capabilities. While some researchers optimistically posit that LLMs possess logical abilities potentially surpassing human performance in certain domains [2], others maintain a more skeptical stance. These critics argue that LLMs fundamentally lack genuine logical reasoning capabilities, suggesting that their inference processes are merely probabilistic approximations, easily disrupted by subtle perturbations [3], [4].

Our research aligns more closely with the latter perspective. However, we hypothesize that it is possible to enhance the logical stability of LLMs during reasoning processes through targeted methodologies. To investigate this hypothesis, we focus on lightweight large language models with fewer than 10 billion parameters. These models, with their relatively limited logical reasoning capabilities, provide an ideal testbed for demonstrating the efficacy of our proposed improvements.

We select mathematical problem-solving as our primary evaluation benchmark, as it presents a rigorous test of a model's reasoning capabilities. Such tasks typically require the decomposition of problems into multiple interdependent steps, each involving precise computation. Correct solutions are only achieved when all reasoning paths are accurately executed. Consequently, mathematical problem-solving presents two fundamental challenges: managing the dependencies between reasoning steps and accurately interpreting conditional text.

The Chain-of-Thought (CoT) [6] approach has emerged as a widely adopted method to address these challenges. By employing prompts such as "Let's think step by step", models can generate detailed reasoning steps, serving as contextual information to significantly improve inference accuracy. However, further investigations have revealed inherent limitations in the CoT method, including sequential errors and comprehension inaccuracies. Models often conflate the logical order of mathematical problems with the textual sequence, and errors in extracting numerical values and computational logic are common [5].

To address these shortcomings, researchers have proposed various methods, including Retrieval-Augmented Generation (RAG) [7] to mitigate hallucination and contrastive learning to optimize logical dependencies [8]. Nevertheless, the efficacy of these methods remains suboptimal when applied to lightweight models. As illustrated in Fig. 1, we observe an interesting phenomenon in mathematical problem similarity: Problem 1(left) and Problem 2(middle) exhibit high semantic similarity. Problem 2 and Problem 3(right) demonstrate strong logical similarity. The algebraic solutions for Problem 1 and Problem 3 are remarkably similar.

This observation leads to a crucial insight: When approaching the solution for Problem 1, it would be more beneficial to use Problem 3 as a reference example due to their shared logical structure and solution approach. Conversely, relying on Problem 2 as a reference, despite its semantic similarity to Problem 1, may introduce incorrect priors and potentially lead to erroneous reasoning paths.

In light of these challenges, we introduce Logic Contrastive Reasoning (LCR) . Our approach introduces a novel method for measuring the similarity of mathematical problems, which we apply to the selection of reference samples in few-shot learning. This method not only considers the semantic similarity of problems but also emphasizes the similarity in reasoning logic. By providing more targeted examples to the model, we achieve improved reasoning accuracy.

Our primary contributions are twofold:

- We propose the concept of logical similarity to retrieve samples with high similarity to the intermediate processes of the problem at hand, thereby standardizing the model's reasoning logic.
- We construct a reasoning pipeline based on Logic Contrastive Reasoning, which significantly enhances the logical reasoning capabilities of lightweight large language models.

Experiments conducted on models with fewer than 7 billion parameters demonstrate that our method achieves SOTA performance in mathematical problem-solving accuracy. These results indicate that even lightweight models, when appropriately augmented, can excel in complex reasoning tasks. Our research opens up promising new avenues for further exploration of the reasoning capabilities of large language models.

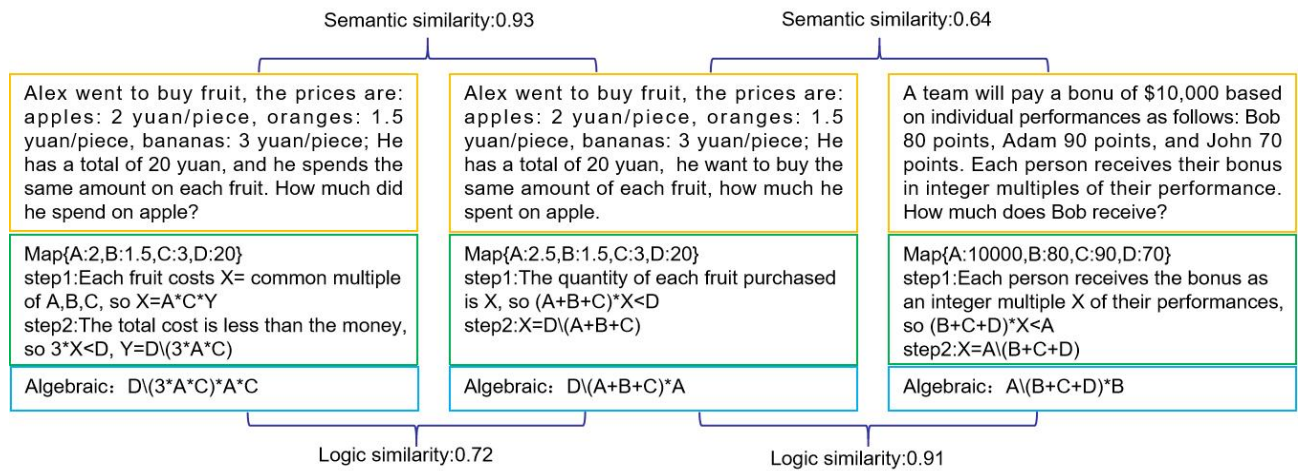

Fig 1. Comparison between semantic similarity and logical similarity metrics in characterizing the similarity of reasoning processes in mathematical problems. For the three mathematical problems shown in the figure, the logical similarity metric more accurately represents the similarity of reasoning processes between problems.

## 2. Related Works

Research on mathematical reasoning using large language models primarily focuses on two main directions: improvements based on prompting techniques and those based on fine-tuning. This article concentrates on prompt-based improvements, which can be further categorized into two approaches: Chain-of-Thought (CoT) methods and self-verification methods.

### 2.1 Cot Prompt for Reasoning

Chain-of-thought (CoT) prompting, introduced by Wei et al. [6], demonstrated that specific prompts like "Let's think step-by-step" can enable language models to perform chain-of-thought reasoning in a zero-shot manner. This breakthrough has inspired numerous works that build upon step-by-step reasoning approaches. For instance, automatic chain-of-thought [12] was proposed to address the challenges associated with manually annotating chain-of-thought demonstrations. Additionally, researchers explored methods to decompose complex problems into multiple sub-problems [23], or even into code programs that can be automatically executed [13]. Contrastive chain-of-thought (CCoT) [11] enhances language model reasoning by proposing a contrastive approach to CoT. Active-Prompt [9] adapted language models to different tasks using task-specific example prompts with manually designed CoT reasoning annotations. To address the crucial question of determining which problems are most important and helpful for annotation, the authors proposed a solution that draws inspiration from uncertainty-based active learning. They introduced several metrics to characterize uncertainty, thereby selecting the most uncertain questions. Faithful CoT [10] approached the problem by separating it into two stages: Translation (Natural Language query → symbolic reasoning chain) and Problem Solving (reasoning chain → answer), utilizing a language model and a deterministic solver, respectively.

### 2.2 Self-verification for LLM

Multiple decoding strategies have been proposed in the literature to improve the output quality of language models. These include temperature sampling, top-k sampling, and minimum Bayes risk decoding [14].

Re-ranking is another common approach to enhance generation quality. Cobbe et al. (2021) [15] demonstrated that training a "verifier" to re-rank generated solutions substantially improves the solve rate on mathematical tasks, compared to merely fine-tuning the language model. Self-correction [18] techniques leverage the law of large numbers by generating multiple answers under identical conditions and selecting the most probable response as the final answer. This method effectively improves answer consistency while simultaneously increasing accuracy, demonstrating that large language models tend to favor correct answers in most cases. Additionally, some approaches incorporate mechanisms of self-critique and self-reflection into large language models, enabling them to refine their outputs. For example, Shinn, Labash et al. (2023) introduced Reflexion [16], a technique that employs external feedback to detect ineffective actions and engage in self-reflection.In self-verification [17], researchers reverse the problem and answer, repeatedly confirming the correctness of the reasoning. This method has also been shown to improve model accuracy.

## 3. Logic Contrastive Reasoning (Method)

Using large language models (LLMs) to solve mathematical application problems presents two main challenges: addressing logical reasoning errors and resolving semantic ambiguities. To address these issues, we draw inspiration from the effective Chain of Thought (CoT) and contrastive methods, and propose a novel approach called Logic Contrastive Reasoning. This method uses a few sample problems as reference examples to guide the model in generating step-by-step solutions, thereby improving solving accuracy.

Logic Contrastive Reasoning comprises two key components: Logic Similarity and Contrastive Reasoning. The first component addresses how to evaluate the similarity of solving processes between two mathematical problems, while the second component focuses on how to incorporate logically similar problems into the prompt to enhance solving accuracy.

### 3.1 Logic Similarity

We use prompts and large language models to logically structure mathematical problems, breaking them down into known conditions and questions to be solved. Through CoT reasoning, we list intermediate questions and solve them step by step to reach the final answer. Each reasoning step can be transformed into an algebraic expression, allowing us to approximate the similarity of reasoning steps through expression similarity.

To represent the complete reasoning process, we combine the individual reasoning steps sequentially, merging the formulas in the solving process into a total solving formula. This transforms the similarity of mathematical problems into a problem of calculating the similarity between two solving formulas. To represent algebraic similarity, we first align variables to eliminate inconsistencies in their order of appearance and position within the formula. We employ a straightforward method of replacing variables in the formula with placeholders. For example, A*(B+C+D)*B is converted to @*(@+@+@)*@.

We then calculate the similarity between algebraic expressions using the Normalized Tree Edit Distance as a metric. This metric measures the proportion of characters that need to be modified for two strings to become identical, expressed as:

$$N(Al_1, Al_2) = \frac{Lev(Al_1, Al_2) + Lev(Al_2, Al_1)}{len(Al_1) + len(Al_2)} \quad (1)$$

where $N$ represents the normalized logical similarity function, $Al$ denotes the algebraic expression after variable alignment, $Lev()$ is the Levenshtein edit distance, and $len$ is the length of the expression string.

Building upon this, the Normalized Tree Edit Distance constructs a tree representation of the formula. However, this approach is highly sensitive to the tree structure. If two mathematical expressions exhibit slight structural differences, even when their mathematical meanings are similar, the calculated edit distance may still be substantial. Furthermore, due to the presence of properties such as commutativity and associativity, there are inherent variations in the ways mathematical expressions can be written, which makes it challenging for the NTED to accurately represent the similarity between expressions. To address this issue, we propose a new metric for measuring expression similarity. First, we divide the expression into two parts by selecting a specific operator outside the parentheses as a splitting point, aiming to create two branches of approximately equal length. This approach allows for a more precise representation of the formula while minimizing excessive branching that could increase structural sensitivity. In this manner, the swapping of branches will not significantly affect the similarity measure. For instance, in Expression 2, when the splitting operator is multiplication or addition, the two branches can be interchanged. Therefore, we take the minimum value of the similarity measures obtained before and after the swap. Conversely, when the splitting operator is division or subtraction, the overall similarity is computed as the sum of the similarities of the two branches in their respective order. The method then calculates the Normalized Edit Distance for each branch and merges them by selecting the minimum value based on branch labels. The expression for the Normalized Tree Edit Distance TD is formulated as follows:

$$TD(Al_{1i}, Al_{2j}) = \begin{cases} \sum_{i=1}^{2} N(Al_{1i}, Al_{2i}), & \text{if } T(Al_1) \notin [+, *] \\ \min(\sum_{i=1}^{2} N(Al_{1i}, Al_{2i}), \sum_{i=1}^{2} N(Al_{1i}, Al_{2(3-i)})), & \text{else} \end{cases}$$

(2)

where $T()$ represents the branch operator which is in $[+,-,\times,\div]$, and min indicates the minimum value.

## 3.2 Contrastive Reasoning

The simplest form of mathematical problem reasoning based on large language models can be represented as $E(Q, A)$, consisting only of Question and Answer components. However, lightweight large models struggle significantly with extracting known conditions from the Question and planning the solution process. Adding prior knowledge to the prompt, including Chain of Thought (CoT) and examples, can improve mathematical reasoning accuracy. This enhanced reasoning can be represented as $E(P, Q, A)$, $P$ is the prompt.

Contrastive Chain of Thought (CCoT) introduces positive and negative examples. These examples not only promote correct derivation by the large model but also help to avoid common errors. The reasoning formula for CCoT is $E(P, Qs, D+, D-, Q, A)$, where $Qs$ represents example questions, D represents the reasoning process, and positive and negative signs indicate correct and incorrect reasoning.

In our approach, we introduce an algebraic solving form, represented by $Al$ for the algebraic reasoning process. Thus, our contrastive reasoning formula becomes $E[P, Qs, D+, Al+, D-, Al-, Q, A]$.

Unlike CCoT's method that synthesizes negative samples, we use math prompts with different models for screening to collect positive and negative examples, **which is a RAG method**. This approach generates negative examples that are more realistic. The process involves:

(1) Logical structuring of the mathematical problem using prompts and large models, breaking it down into known conditions and questions to be solved (Step 1 in Algorithm).

(2) Identifying intermediate questions in the logical reasoning based on known conditions and the problem to be solved. These are represented in text form, followed by mathematical reasoning to obtain answers (Step 2 in Table 1).

(3) Expressing the reasoning process in algebraic form, yielding algebraic expressions and solutions (Step 3 in Table 1).

Combining text-form reasoning and algebraic reasoning to deduce the final answer.

Then we solve each problem multiple times and compare the solutions with the true values to identify which problems are answered correctly or incorrectly. Samples with both correct and incorrect answers for the same problem are included in the example sample set. Based on this sample set, we use the logic similarity function to retrieve several examples relevant to the problem being solved, as shown in Step 5 of Table 1. Finally, we construct a contrastive reasoning prompt, as illustrated in Prompt 4 of Table 1, to solve the mathematical problem. The $TLS()$ function in the algorithm is an integrated function that combines both semantic and logical aspects. In the context of retrieving similar mathematical problems, both semantic information and logical information play important roles. Therefore, based on the logical similarity of the formulas, we incorporate semantic similarity by using hyperparameters to combine these two similarity measures, thereby representing the overall similarity between two mathematical problems.

As shown in Equation 3, $TD()$ denotes the similarity between mathematical expressions $Al_1$ and $Al_2$ as in Equation 2. The $Sem()$ function utilizes the semantic similarity model SentBERT[19] for its computation. SentBERT is specifically designed to assess sentence-level semantic similarity. In our approach, we first encode the mathematical problems, $Q_1$ and $Q_2$, into high-dimensional vector representations using SentBERT. This model captures contextual information and semantic relationships within the expressions by leveraging a transformer-based architecture. Once the encoding is complete, we calculate the cosine similarity between the resulting vectors. This cosine similarity score reflects how closely related the two mathematical problems are in terms of their underlying semantic content. Finally, we set the hyper-parameter empirically at a value of 0.7.

$$TLS(ex_1, ex_2) = \alpha TD(Al_1, Al_2) + \bar{\alpha} Sem(Q_1, Q_2) \quad (3)$$
$$ex_1 = (Q_1, Al_1)$$

where $\bar{\alpha} = 1 - \alpha$

Algorithm. Pipeline for LCR

Sample Preprocessing

Step 1: Extract Known Conditions from $Q$ Using a Prompt

Prompt 1: "List the known conditions."

Step 2: Develop a Reasoning Process Based on Known Conditions and the Problem to be Solved, then Solve the Problem Step by Step

Prompt 2: "Let's first understand the problem and devise a plan to solve it. Then, let's carry out the plan to solve the problem step by step."

Step 3: Convert Known Conditions and Solution Process into Algebraic Form

Prompt 3: "Transform the conditions into algebraic form using a key-value mapping. Then, convert the solving steps into algebraic form."

Step 4: Summarize the Above Process to Obtain the Solution to the Problem

Contrastive reasoning

Step 5: Use Logical Similarity Function to Retrieve Similar Samples

We use the Logical Similarity Function $TLS()$ to retrieve samples from the example set that are similar to the problem to be solved:

$\{ex_1, ex_2\} = sorted(TLS(ex_0, ex_1), TLS(ex_0, ex_2), ..., TLS(ex_0, ex_n))$

Where: $ex_i$ represents the i-th example sample, including $Qs, D+, AI+, D-, AI-, ex_0$ is the problem to be solved, $\{\}$ contains the $ex_j$ samples that are calculated to be most logically similar to the problem to be solved

$sorted()$ is a function that sorts the values in descending order

Step 6: Construct Reasoning Prompt and Solve the Problem

Prompt4: Given a math problem, please solve it step by step. Please follow the examples $\{ex_1, ex_2\}$.

## 4. Experiments

### 4.1 Experimental Setup

To validate the effectiveness of our proposed logic similarity-based reasoning method for lightweight large language models, we conduct evaluations on commonly used mathematical question-answering datasets. We utilize GSM8K [15] and SVAMP [22] datasets, both of which include problem descriptions and answers. SVAMP contains relatively simple problems with direct answers, while GSM8K presents more complex problems with complete reasoning processes in the answers.

Our experiments employ the following lightweight models: Mistral-7B [20], LLaMA2-7B [21]. These models are loaded using 4-bit quantization (INT4) to optimize memory usage. Inference parameters are set with a generation length of 400, top_p of 0.95, temperature of 0.1, top_k of 30, and a repetition penalty of 1.15. To demonstrate the method's efficacy across model scales, we also validate our approach on a 175B parameter model, specifically using the ChatGPT-3.5 Turbo 0301 4K version.

We use accuracy as our primary evaluation metric. To ensure fairness in our experiments, we utilize 100 test samples from both SVAMP and GSM8K datasets, conducting multiple test runs for robust results.

### 4.2 Main Results

We design multiple experiments to demonstrate the effectiveness of our proposed method. The process involved two main steps:

(1) Sample Selection: As shown in Table 1, we first screen for positive and negative examples. To reduce the computational cost of using the entire training set as reference samples, we apply logical similarity filtering to the training set. This process eliminate samples with high similarity.

(2) Mathematical Problem Solving: Using the select positive and negative examples, we conduct mathematical problem-solving experiments, comparing various parameter settings and algorithms.

#### 4.2.1 Parameter Optimization

Using the Mistral-7B model, we evaluate the impact of two key parameters: the number of few-shot examples and the number of guess attempts. We test few-shot examples ranging from 1 to 9 (odd numbers only) and guess attempts of 5 and 10.

Fig. 2 illustrates our findings. The horizontal axis represents the number of few-shot examples, while the vertical axis denotes the prediction accuracy. "Guess5"

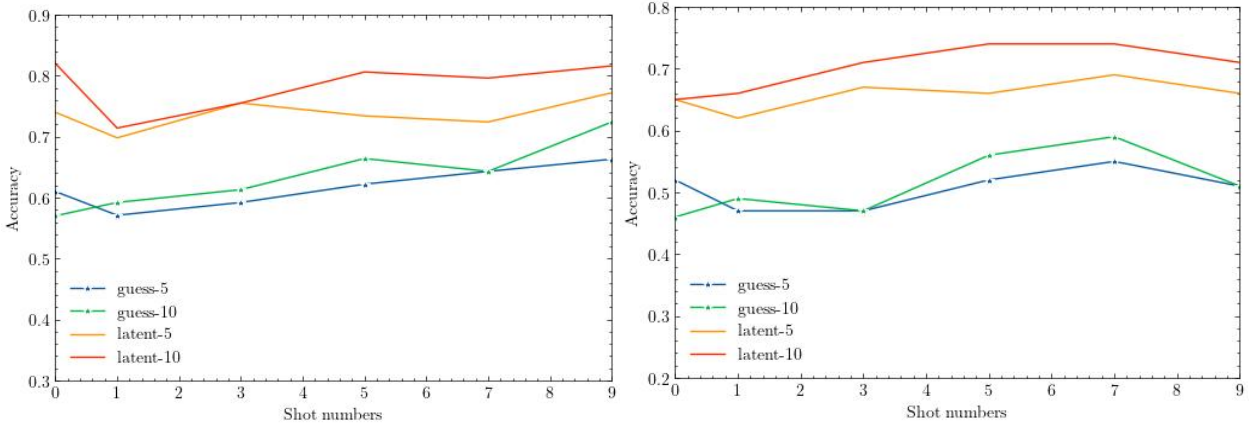

Fig 2. Accuracy Curves Under Different LCR Parameters. For the SVAMP dataset (left graph),the highest test accuracy is achieved with 10 guess attempts and 9 reference samples. For the GSM8K dataset (right graph),the peak accuracy of 59.2% is reached with 10 guess attempts and 7 reference samples.

and "Guess10" represent different numbers of prediction attempts. Our results demonstrate that:

(1) Increasing the number of guess attempts generally improves prediction accuracy.

(2) A higher number of few-shot examples correlates with improved prediction accuracy.

(3) The "Latent" accuracy, representing the model's potential accuracy when any of the multiple predictions is correct, can be viewed as an upper bound of the model's performance. This suggests that correct reasoning paths exist within the model, but probabilistic factors may hinder their consistent use.

Fig. 2 presents results for the GSM8K dataset under similar experimental conditions. We observe that while a higher number of guess attempts increase the latent accuracy, the actual accuracy change slightly when the guesses are increased from 5 to 10. Regarding the few-shot parameter, performance peak at 7 examples and declined at 9, indicating a threshold effect in the utility of reference samples for accuracy improvement.These results provide valuable insights into the optimal configuration of our proposed method and demonstrate its effectiveness across different problem complexities and model scales.

We compared our method with state-of-the-art algorithms: Chain-of-Thought (CoT), Self-Correction(SC), and Plan-and-Solve(PS). Table 2 summarizes the results on SVAMP and GSM8K datasets using the Mistral-7B model. Our method consistently outperforms existing approaches on both datasets, demonstrating significant improvements in prediction accuracy. We achieve a 15.8% and 18.6% improvement over CoT On SVAMP and GSM8K respectively. Indeed, the SC method significantly contribute to accuracy improvement. Without SC, our LRC method still outperform the CoT method, increasing accuracy by 7.6% on SVAMP and 7.4% on GSM8K respectively. These results highlight the effectiveness of our logic similarity-based reasoning method in enhancing mathematical problem-solving capabilities of lightweight large language models across varying problem complexities.

Table 2. Accuracy comparison on GSM8K and SVAMP datasets with 5 methods. The best results are boldfaced.

|  | Gsm8k | | SVAMP | |
|---|---|---|---|---|
|  | Accuracy | Latent acc | Accuracy | Latent acc |
| CoT[6] | 37.7 | / | 56.6 | / |
| SC[18] | 47.0 | 60.2 | 59.5 | 82.3 |
| PS[24] | 50.5 | 63.1 | 62.8 | 84.9 |
| LCR(woSC) | 45.1 | / | 64.2 | / |
| LCR(wSC) | **59.2** | 71.6 | **72.4** | 84.1 |

### 4.2.2 Contrastive Learning Experiments

To validate the effectiveness of our proposed logical similarity retrieval, we conduct experiments comparing various contrastive learning strategies. We employ two methods for selecting example samples: (1) Random

Selection: A fixed number of example samples are randomly selected, potentially unrelated to the inference sample, and use consistently for each inference. (2) Similarity-based Selection: Example samples are retrieved based on their similarity (semantic or logical) to the current mathematical problem, varying for each inference.

Random selection methods include Fix, Hard, and Contrastive CoT (see Appendix), while others use similarity-based selection. According to Table 3, on SVAMP, semantic similarity selection has a negative impact, likely due to samples differing only in numerical values or solution objects, leading to misleading guidance. On GSM8K, semantic similarity perform comparably to random selection. Fixed example samples are effective, especially Contrastive SC, which include positive and negative examples and error analysis, showing the most significant improvement in accuracy. These findings demonstrate the superiority of our logical similarity-based approach over semantic similarity or random selection methods in enhancing mathematical problem-solving capabilities.

Table 3 Accuracy Comparison with 5 comparative learning strategies.

|  | SVAMP | | Gsm8k | |
| --- | --- | --- | --- | --- |
|  | guess5 | guess10 | Guess5 | Guess10 |
| Fix | 62.1 | 70.3 | 43.7 | 48.7 |
| Hard | 62.9 | 70.9 | 50.0 | 55.1 |
| Contrastive | 63.4 | 71.5 | 53.2 | **59.4** |
| Semantic RAG | 60.2 | 65.5 | 51.1 | 54.5 |
| LCR(ours) | **66.7** | **72.4** | **55.1** | 59.2 |
| Logic RAG | 63.6 | 70.8 | 52.6 | 57.9 |

In terms of consistency, the use of RAG (Retrieval-Augmented Generation) has significantly improved the consistency of model predictions. Specifically, models that utilize RAG demonstrate greater stability in generating results, whereas models without RAG tend to produce a variety of inconsistent predictions. This inconsistency increases the risk of erroneous predictions, as the model may provide drastically different answers in different contexts. Therefore, the introduction of RAG not only helps improve prediction accuracy but also reduces the likelihood of errors caused by the diversity of model outputs.

### 4.2.3 Error analysis

We conducted an analysis and statistical study of common errors in two datasets to understand how lightweight large language models might make mistakes in mathematical reasoning. We defined four types of errors: comprehension errors, calculation errors, logical errors, and formula errors. Comprehension errors occur when the model misunderstands the object to be solved. For example: Question: Zachary did 53 push-ups and 14 crunches, whereas David did 17 more push-ups but 10 fewer crunches than Zachary. How many push-ups and crunches did David do? A comprehension error would occur if the model misinterpreted the question and solved for Zachary's exercises instead of David's. Calculation errors involve incorrect arithmetic results. For instance, incorrectly calculating 244 * 146 = 35,232 (the correct answer is 35,624).Formula errors occur when there's an inconsistency between the text and the formula, or when formula derivation is inconsistent. For example, "Among the 200 Grade 5 students, 2/5 are boys. So, there are 200 * (1 - 2/5) = 200 * (3/5) = 120 boys". Logical errors happen when the reasoning is confused, leading to a solution that doesn't match the original problem. For instance, "To find out how many more books than action figures are on Jerry's shelf, we simply subtract the number of action figures from the total number of items: 12 (total items) - 5 (action figures) = 7."

From a statistical perspective, comprehension errors and logical errors were the most prevalent, while calculation errors and formula errors occurred less frequently. In Table 4, the error distributions in two datasets are shown, the proportion of the major errors are 64.3% and 73.8%. Given these statistics, the most critical area for improvement is enhancing the model's logical reasoning capabilities. Two potential approaches to address this are: (1) Pre-training approach: Expand the pre-training dataset to include more mathematics, coding, and related content. This method aims to improve the model's foundational understanding of logical and mathematical concepts. (2) Fine-tuning approach: During the supervised fine-tuning (SFT) stage, incorporate multi-step reasoning data. This approach focuses on teaching the model how to break down complex problems into manageable steps. In comparison, the second approach (fine-tuning with multi-step reasoning data) is likely to be more feasible and cost-effective in terms of resources required.

Table 4. Reasoning error analysis on SVAMP and GSM8K.

|  | SVAMP | GSM8k |
|---|---|---|
| Comprehension error | 10(35.7%) | 14(33.3%) |
| Calculation error | 6(21.4%) | 8(19%) |
| Logic error | 8(28.6%) | 17(40.5%) |
| Equation error | 4(14.3%) | 3(7%) |
| Total | 28 | 42 |

#### 4.2.4 Generalization experiments

The proposed method is not only applicable to lightweight models but can also be extended to large-scale language models. To demonstrate this, we conducted tests using ChatGPT-3.5 (turbo0314 version) on the GSM8K and SVAMP datasets. The parameters were set to 7 reference samples and 5 guess attempts. The results presented in Table 5 show that: On the GSM8K dataset, LCR + ChatGPT (without SC) outperforms the baseline CoT + ChatGPT method by 12.0 %. LCR + ChatGPT with the SC strategy shows a 12.5 % improvement over the SC + PaLM combination on the GSM8K dataset. However, on the SVAMP dataset, our method slightly underperforms compared to SC + PaLM. The relatively lower performance on SVAMP can be attributed to the high similarity between some questions in the reference set and the test set. In these cases, the reference samples may mislead the large language model in understanding and reasoning about the problem, leading to incorrect solutions.

Table 5 Performance comparison on GSM8K and SVAMP datasets with various model sizes and prompts.

|  | Gsm8k | SVAMP | Model size |
|---|---|---|---|
| Mistral+LCR(ours) | 59.2 | 72.4 | 7b |
| CoT+Chatgpt[6] | 61.7 | 77.6 | 175b |
| Self verification[17] | 65.1 | 76.9 | 175b |
| Active prompt(chat)[9] | 65.6 | 80.5 | 175b |
| PS+Chatgpt[24] | 70.7 | 81.7 | 175b |
| SC+Palm[18] | 74.4 | 86.6 | 540b |
| Faithful Cot[10] | 80.0 | **88.8** | 175b |
| Contrastive CoT[11] | 86.2 | 85.2 | 175b |
| Chatgpt+LCR(woSC) | 73.7 | 81.0 | 175b |
| Chatgpt+LCR | **86.9** | 84.3 | 175b |

## 5. Conclusion

To address the challenges of hallucinations and logical errors in mathematical reasoning tasks for lightweight large language models, we proposed a novel approach leveraging contrastive learning. Our approach enabled the automatic selection of a set of reference problems that shared logical similarities with the target problem. Using these positive and negative examples, we constructed tailored prompts that guide the language model to adopt reasoning strategies similar to the positive examples while avoiding errors common to the negative ones.

Experiments conducted on multiple public mathematical problem datasets demonstrated significant improvements over existing state-of-the-art methods. Furthermore, we successfully extended this method to a large language model with 175 billion parameters, achieving results comparable to optimal human performance on both datasets. Finally, we provided a comprehensive analysis of common issues encountered during the reasoning process of large language models. This analysis offers valuable insights for future enhancements in the reasoning capabilities of these models.

**Declaration of generative AI and AI-assisted technologies in the writing process**

## References

[1]. Floridi L, Chiriatti M. **GPT-3**: Its nature, scope, limits, and consequences[J]. Minds and Machines, 2020, 30: 681-694.

[2]. Bubeck, S., Chandrasekaran, V., Eldan, R., Gehrke, J.,Horvitz, E., Kamar, E., Lee, P., Lee, Y. T., Li, Y.,Lundberg, S., et al. **Sparks** of artificial general intelligence: Early experiments with gpt-4. arXiv preprint arXiv:2303.12712, 2023.

[3]. Dziri, N., Lu, X., Sclar, M., Li, X. L., Jiang, L., Lin, B. Y., Welleck, S., West, P., Bhagavatula, C., Bras, R. L., Hwang, J. D., Sanyal, S., Ren, X., Ettinger, A., Harchaoui, Z., and Choi, Y. **Faith** and fate: Limits of transformers on compositionality. In Thirty-seventh Conference on Neural Information Processing Systems, 2023.

[4]. Verma, M., Bhambri, S., and Kambhampati, S. Theory of mind abilities of large language models in human-robot interaction: **An illusion**? arXiv preprint arXiv:2401.05302, 2024b.

[5]. McCoy, R. T., Yao, S., Friedman, D., Hardy, M., and Griffiths, T. L. **Embers of auto regression**: Understanding large language models through the problem they are trained to solve. arXiv preprint arXiv:2309.13638, 2023

[6]. Wei J, Wang X, Schuurmans D, et al. **Chain-of-thought** prompting elicits reasoning in large language models[J]. Advances in neural information processing systems, 2022, 35: 24824-24837.

[7]. Lewis P, Perez E, Piktus A, et al. **Retrieval-augmented generation** for knowledge-intensive nlp tasks[J]. Advances in Neural Information Processing Systems, 2020, 33: 9459-9474.

[8]. Cot-p Boshi Wang, Sewon Min, Xiang Deng, Jiaming Shen,You Wu, Luke Zettlemoyer, and Huan Sun. 2023.Towards **understanding chain-of-thought** prompting:An empirical study of what matters. In Proceedings of the 61st Annual Meeting of the Association for Computational Linguistics (Volume 1: Long Papers),pages 2717–2739, Toronto, Canada.

[9]. Diao, S., Wang, P., Lin, Y., & Zhang, T. (2023). **Active prompting** with chain-of-thought for large language models. *arXiv preprint arXiv:2302.12246.*

[10]. Lyu, Q., Havaldar, S., Stein, A., Zhang, L., Rao, D., Wong, E., ... & Callison-Burch, C. (2023). **Faithful chain-of-**thought reasoning. *arXiv preprint arXiv:2301.13379.*

[11]. Chia, Y. K., Chen, G., Tuan, L. A., Poria, S., & Bing, L. (2023). **Contrastive chain-of-thought** prompting. *arXiv preprint arXiv:2311.09277.*

[12]. Zhuosheng Zhang, Aston Zhang, Mu Li, and Alex Smola. 2023. **Automatic chain of thought** prompting in large language models. In The Eleventh International Conference on Learning Representations.

[13]. Program-cot Luyu Gao, Aman Madaan, Shuyan Zhou, Uri Alon, Pengfei Liu, Yiming Yang, Jamie Callan, and Graham Neubig. 2023. **PAL**: Program-aided language models. In Proceedings of the 40th International Conference on Machine Learning, volume 202 of Proceedings of Machine Learning Research, pages 10764–10799. PMLR

[14]. Freda Shi, Daniel Fried, Marjan Ghazvininejad, Luke Zettlemoyer, and Sida I. Wang. Natural language to code translation with execution. In Proceedings of the 2022 Conference on Empirical Methods in Natural Language Processing, pp. 3533–3546,December 2022.

[15]. T-verify/Gsm8k Karl Cobbe, Vineet Kosaraju, Mohammad Bavarian, Mark Chen, Heewoo Jun, Lukasz Kaiser,Matthias Plappert, Jerry Tworek, Jacob Hilton, Reiichiro Nakano, Christopher Hesse, and John Schulman. **Training verifiers** to solve math word problems, 2021

[16]. Shinn N, Labash B, Gopinath A. Reflexion: an autonomous agent with dynamic memory and **self-reflection**[J]. arXiv preprint arXiv:2303.11366, 2023.

[17]. Weng Y, Zhu M, Xia F, et al. Large Language Models are Better Reasoners with **Self-Verification**[C]//The 2023 Conference on Empirical Methods in Natural Language Processing. 2023.

[18]. Wang X, Wei J, Schuurmans D, et al. **Self-consistency** improves chain of thought reasoning in language models[J]. arXiv preprint arXiv:2203.11171, 2022.

[19]. Reimers N, Gurevych I. **Sentence-BERT**: Sentence Embeddings using Siamese BERT-Networks[C]//Proceedings of the 2019 Conference on Empirical Methods in Natural Language Processing and the 9th International Joint Conference on Natural Language Processing (EMNLP-IJCNLP). 2019: 3982-3992.

[20]. Jiang A Q, Sablayrolles A, Mensch A, et al. **Mistral** 7B[J]. arXiv preprint arXiv:2310.06825, 2023.

[21]. Touvron H, Martin L, Stone K, et al. Llama 2: Open foundation and fine-tuned chat models[J]. arXiv preprint arXiv:2307.09288, 2023.

[22]. Svamp Arkil Patel, Satwik Bhattamishra, and Navin Goyal. Are NLP models really able to solve simple math word problems? In Proceedings of the 2021 Conference of the North American Chapter of the Association for Computational Linguistics: Human Language Technologies, pp. 2080–2094, Online, June 2021.

[23]. Lei Wang, Wanyu Xu, Yihuai Lan, Zhiqiang Hu, Yunshi Lan, Roy Ka-Wei Lee, Ee-Peng Lim, Plan-and-Solve Prompting: Improving Zero-Shot Chain-of-Thought Reasoning by Large Language Models, arXiv:2305.04091v3, 2023

**Appendix**

**Fix_example prompt:**

These are some examples for solving math problem:

Q: There are 15 trees in the grove. Grove workers will plant trees in the grove today. After they are done,
there will be 21 trees. How many trees did the grove workers plant today?

A: We start with 15 trees. Later we have 21 trees. The difference must be the number of trees they planted.
So, they must have planted 21 - 15 = 6 trees. The answer is 6.

Q: If there are 3 cars in the parking lot and 2 more cars arrive, how many cars are in the parking lot?

A: There are 3 cars in the parking lot already. 2 more arrive. Now there are 3 + 2 = 5 cars. The answer is 5.

Q: Leah had 32 chocolates and her sister had 42. If they ate 35, how many pieces do they have left in total?

A: Leah had 32 chocolates and Leah's sister had 42. That means there were originally 32 + 42 = 74
chocolates. 35 have been eaten. So in total they still have 74 - 35 = 39 chocolates. The answer is 39.

Q: Jason had 20 lollipops. He gave Denny some lollipops. Now Jason has 12 lollipops. How many lollipops
did Jason give to Denny?

A: Jason had 20 lollipops. Since he only has 12 now, he must have given the rest to Denny. The number of
lollipops he has given to Denny must have been 20 - 12 = 8 lollipops. The answer is 8.

Q: Shawn has five toys. For Christmas, he got two toys each from his mom and dad. How many toys does he
have now?

A: He has 5 toys. He got 2 from mom, so after that he has 5 + 2 = 7 toys. Then he got 2 more from dad, so
in total he has 7 + 2 = 9 toys. The answer is 9.

Q: There were nine computers in the server room. Five more computers were installed each day, from
monday to thursday. How many computers are now in the server room?

A: There are 4 days from monday to thursday. 5 computers were added each day. That means in total 4 * 5 =
20 computers were added. There were 9 computers in the beginning, so now there are 9 + 20 = 29 computers.
The answer is 29.

Q: Michael had 58 golf balls. On tuesday, he lost 23 golf balls. On wednesday, he lost 2 more. How many
golf balls did he have at the end of wednesday?

A: Michael initially had 58 balls. He lost 23 on Tuesday, so after that he has 58 - 23 = 35 balls. On

Wednesday he lost 2 more so now he has 35 - 2 = 33 balls. The answer is 33.

Q: Olivia has $23. She bought five bagels for $3 each. How much money does she have left?

A: She bought 5 bagels for $3 each. This means she spent 5 * $3 = $15 on the bagels. She had $23 in

beginning, so now she has $23 - $15 = $8. The answer is 8.

**Hard_example prompt:**

These are some examples for solving math problem:

Q:Frank was reading through his favorite book. The book had 2 chapters each with 405 pages. It took frank 664 days to finish the book. How many days did it token for reading one chapter?

A: the answer is 664.0 / 2.0, so it is 332.0

Q:The Razorback t-shirt shop sells each t-shirt for $ 51 dollars. During the Arkansas and Texas tech game they offered a discount of $ 8 per t-shirt and sold 130 t-shirts. How much money did they make from selling the t-shirts?

A:t-shirt price with discount is 51.0 - 8.0, they sell 130 t-shirts, so totally they make ( 51.0 - 8.0 ) * 130.0 dollars, therefor the answer is $5590.0.

Q:Paige raised 7 goldfish and 12 catfish in the pond but stray cats loved eating them. Now she has 15 left.How many fishes disappeared?

A:Paige raised 7.0 + 12.0 fishes in the pond, now she has 15 left, so the cats eat ( 7.0 + 12.0 ) - 15.0 fishes, therefor the answer is 4.0 fishes.

Q:Faye had 35 packs of pencils each one having 4 pencils. She was placing her pencils into rows with 2 pencils in each row. How many rows could she make?

A:Faye had 35.0 * 4.0 pencils in all, 2 pencils in each row, so there are 35.0 * 4.0 / 2.0 rows, therefore the answer is 70.0.

Q:The Razorback shop makes $ 192 dollars off each t-shirt and $ 34 off each jersey. During the Arkansas and Texas tech game they sold 157 t-shirts and 19 jerseys. How much more does a t-shirt cost than a jersey?

A:the shop makes $192.0 each t-shirt, so the t-shirt price is $192.0, then we can see the jersey price is 34.0, therefor t-shirt cost $192.0 - $34.0 more, the answer is 158.0.

Q:6 birds and 3 storks were sitting on the fence. 2 more storks came to join them.How many more birds than storks are sitting on the fence?

A:6 birds and 3 storks were sitting on the fence,2 more storks came , then there were 3.0+2.0 storks, so birds are 6.0 - ( 3.0 + 2.0 ) more than storks. the answer is 1.0

Q:Paul had 2 books. After selling some in a garage sale he bought 150 new ones. If he has 58 books now How many books did he sell?

A:Paul had 2 books, then he bought 150 new ones, so he should have ( 2.0 + 150.0 ) books, but he sell some, then he had 58.0 books, so he selled ( 2.0 + 150.0 ) - 58.0 books. the answer is 94.0.

Q:Each Ferris wheel in paradise park has 19 seats. Each seat in a Ferris wheel can hold 15 people.How many people can ride 20 Ferris wheels at the same time?

A:One Ferris wheel has 19 seats, one seat hold 15 people, so one Ferris wheel can hold 19.0 * 15.0 people. there are 20 Ferris wheels  then ( 19.0 * 15.0 ) * 20.0 people can ride on the same time. the answer is 5700.

**Contrastive prompt:**

The following are examples of math problems and their solutions, which have wrong and right answers.  Please refer to the right answer to solve the new problem, and also avoid mistakes in the wrong answers.

Question1: Leah had 32 chocolates and her sister had 42.If they ate 35, how many pieces do they have left in total?

 Right Answer: Originally, Leah had 32 chocolates and her sister had 42. So in total they had 32 + 42 = 74. After eating 35, they had 74 - 35 = 39 pieces left in total.

Wrong Answer: Originally, Leah had 32 + 42 = 74 chocolates and her sister had 32. So in total they had 74 - 35 = 39. After eating 35, they had 42 pieces left in total.

Explanation: The reason this answer is incorrect is because it states that Leah's sister had 32 chocolates,

which is wrong. The question clearly mentions that Leah's sister had 42 chocolates. |EOS|

Question2:Frank was reading through his favorite book. The book had 2 chapters each with 405 pages. It took frank 664 days to finish the book. How many days did it token for reading one chapter?

Right Answer: The book had 2 chapters, and each chapter had 405 pages. It took Frank 664 days to finish the entire book. Since there were 2 chapters, we can divide the total number of days (664) by the number of chapters (2) to find the number of days it took to read one chapter. Therefore, it took Frank 664 days / 2 chapters = 332 days to read one chapter.

Wrong Answer: The book had 2 chapters, and each chapter had 405 pages. So the total number of pages in the book was 810 pages. It took Frank 664 days to finish the entire book. Since the book had 810 pages, and Frank took 664 days to finish it, he must have read 810 pages in 664 days. Therefore, it took Frank 664 days to read one chapter.

Explanation: In the wrong answer, the critical mistake is that Frank read the entire book of 810 pages in 664 days . However, this contradicts the given information that the book had 2 chapters, and Frank finished the entire book in 664 days. |EOS|

Question3:The Razorback t-shirt shop sells each t-shirt for $ 51 dollars. During the Arkansas and Texas tech game they offered a discount of $ 8 per t-shirt and sold 130 t-shirts. How much money did they make from selling the t-shirts?

Right Answer:t-shirt price with discount is 51.0 - 8.0, they sell 130 t-shirts, so totally they make ( 51.0 - 8.0 ) * 130.0 dollars, therefor the answer is $5590.0.

Wrong Answer: The Razorback t-shirt shop sells each t-shirt for $51. During the Arkansas and Texas Tech game, they offered a discount of $8 per t-shirt and sold 130 t-shirts. Therefore, the total money they made is $51 × 130 = $6,630, as they sold each t-shirt for the original price of $51.

Explanation: In the wrong answer, the total money they made is $51 x 130 = $6,630. This is incorrect because it does not consider the $8 discount offered during the game. |EOS|

Question4:Paige raised 7 goldfish and 12 catfish in the pond but stray cats loved eating them. Now she has 15 left. How many fishes disappeared?

Right Answer: Paige raised 7.0 + 12.0 fishes in the pond, now she has 15 left, so the cats eat ( 7.0 + 12.0 ) - 15.0 fishes, therefor the answer is 4.0 fishes.

Wrong Answer: Initially, Paige had a total of 7 goldfish + 12 catfish = 19 fish. Now, she has 15 fish left, which means 19 - 15 = 4 fish are left. Therefore, 19 fish disappeared from Paige's pond.

Explanation: The wrong answer incorrectly assumes that the remaining 15 fish are the ones that disappeared, instead of the ones that are still left in the pond. The correct approach is to subtract the remaining fish 15 from the initial number of fish 19 to find the number of fish that disappeared 4. |EOS|